\documentclass[conference]{IEEEtran}
\IEEEoverridecommandlockouts
\usepackage{cite}
\usepackage{amsmath,amssymb,amsfonts}
\usepackage{algorithmic}
\usepackage{graphicx}
\usepackage{textcomp}
\usepackage{subcaption}
\usepackage{xcolor}
\usepackage{algorithm2e}

\def\BibTeX{{\rm B\kern-.05em{\sc i\kern-.025em b}\kern-.08em
    T\kern-.1667em\lower.7ex\hbox{E}\kern-.125emX}}
\begin{document}

\title{Policy Based Inference in Trick-Taking Card Games\\
\thanks{
}
}

\author{\IEEEauthorblockN{Douglas Rebstock\IEEEauthorrefmark{1},
Christopher Solinas\IEEEauthorrefmark{2}, Michael Buro\IEEEauthorrefmark{3} and
Nathan R. Sturtevant\IEEEauthorrefmark{4}}
\IEEEauthorblockA{Department of Computing Science \\
University of Alberta\\
Edmonton, Canada\\
Email: \IEEEauthorrefmark{1}drebstoc@ualberta.ca,
\IEEEauthorrefmark{2}solinas@ualberta.ca,
\IEEEauthorrefmark{3}mburo@ualberta.ca,
\IEEEauthorrefmark{4}nathanst@ualberta.ca}}

\maketitle

{\let\thefootnote\relax\footnote{\textcopyright 2019 IEEE.  Personal use of this material is permitted.  Permission from IEEE must be obtained for all other uses, in any current or future media, including reprinting/republishing this material for advertising or promotional purposes, creating new collective works, for resale or redistribution to servers or lists, or reuse of any copyrighted component of this work in other works.}

\begin{abstract}

Trick-taking card games feature a large amount of private information that
slowly gets revealed through a long sequence of actions. This makes the number
of histories exponentially large in the action sequence length, as well as
creating extremely large information sets. As a result, these games become too large to solve. To deal
with these issues many algorithms employ inference, the estimation of the
probability of states within an information set. In this paper, we demonstrate
a Policy Based Inference (PI) algorithm that uses player modelling to infer
the probability we are in a given state. We perform experiments
in the German trick-taking card game Skat, in which we show that this method
vastly improves the inference as compared to previous work, and increases the
performance of the state-of-the-art Skat AI system Kermit when it
is employed into its determinized search algorithm.

\end{abstract}


\section{Introduction} \label{sec:intro}

Determinized search algorithms allow for the application of perfect
information algorithms to imperfect information games.  While this may not
always be a good idea, in some cases it represents the current
state-of-the-art.  These algorithms are composed of two steps: sampling and
evaluation.  First, a state is sampled from the player's current information
set; informally, an information set is a set of states that a player cannot
tell apart given their observations.  After a state
is sampled, it is evaluated using a perfect information algorithm such as
minimax.

Inference is a central concept in imperfect information games.  It involves
using a model of the opponent's play to determine their hidden
information based on the actions taken in the game so far.  Because the states
that constitute the player's information set are not always equally likely,
inference plays a key role in the performance of determinized search
algorithms.

While counter-factual regret (CFR) techniques have produced super-human AI in Poker \cite{moravvcik2017deepstack,brown2018superhuman}, they have not proven useful for trick based games. This is due to the extremely large size of the information sets, the
long length of bidding and cardplay sequences, and the difficulty in creating expressive abstractions. While these long sequences make
the game too large to solve, they also
slowly reveal the private information of the other players, thus making
inference a desirable approach. 

In this paper, we show how an opponent model can be used for inference in
trick-taking card games.  In particular, we train policies on supervised human
data and use them to infer the private information of opponents and partners based on each of their previous actions.  This leads to
improvements over the previous state-of-the-art techniques for inference in
the domain of Skat.

The rest of this paper is organized as follows.  First, we explain the basic
rules of Skat and then work related to inference in
trick-taking card games.  Next, we outline an algorithm for performing
inference in trick-taking card games using an opponent model trained on data
from a diverse set of human players which we term Policy Inference (PI). This
algorithm assumes a policy of the opponents, and directly estimates the reach
probability of a sampled state by computing the product of all probabilities
of the actions in the history given that sampled state.  We evaluate this
algorithm empirically in Skat and show that it significantly outperforms
previous work both in tournament settings and at selecting the true underlying
state. Finally, we conclude the paper and provide ideas for future research.

\section{Background} \label{sec:bg}

Trick-taking card games, like Contract Bridge, Skat, and Hearts,  are imperfect
information games in which information set sizes shrink rapidly due to hidden 
information being revealed by player actions.
Long et al.~\cite{long2010understanding} explain why this is an appropriate 
setting for determinized search algorithms such as Perfect Information Monte
Carlo \cite{levy1989million} and Information Set Monte Carlo Tree Search 
\cite{cowling2012information}.
These algorithms are considered state-of-the-art in several trick-taking card
games, including Bridge \cite{ginsberg2001gib} and 
Skat \cite{furtak2013recursive}.

After sampling, states are evaluated using perfect information evaluation
techniques, but this can be problematic.
In perfect information game trees, the values of nodes depend only on the
values of their children, but in imperfect information games, a node's value
can depend on other parts of the tree.
This issue, called non-locality, is one of the main reasons why determinized
search has been heavily criticized in prior work 
\cite{frank1998search,russell2016artificial}.
Inference helps with non-locality by biasing state samples so that they are
more realistic with respect to the actions that the opponent has made.
This seems to improve the overall performance of determinized algorithms.
However, the gains provided by inference come at the cost of increasing 
the player's exploitability.
If the inference model is incorrect or has been deceived by a clever opponent,
using it can result in low-quality play against specific opponents.

\subsection{Related Work}

Previous applications of determinized search in trick-taking card games
acknowledge the relationship between inference and playing performance.  The
first successful application of determinized search in a trick-taking card
game was GIB \cite{ginsberg2001gib} in Contract Bridge.  The author suggests
that only deals consistent with the actions taken so far are sampled for
evaluation.  More specific details are not provided.
WBridge5 \cite{wbridge5} and Jack \cite{jackbridge} have had recent success in
the World Computer Bridge Championship \cite{bridgebotchampionship}, but their
implementation details are not readily available.

In Skat, Kermit \cite{buro2009improving,furtak2013recursive} used a
table-based technique to bias state sampling based on opponent bids and
declarations.  This approach only accounts for a limited amount of the
available state information and neglects important inference opportunities
that occur when opponents play specific cards. This inference will be referred to as Kermit Inference (KI) for the rest of the paper. 
Solinas et al.~\cite{solinas2019improving} extend this process by using a neural 
network to make predictions about individual card locations. By assuming 
independence between these predictions, the probability of a given configuration
was calculated by multiplying the probabilities corresponding to card locations
in the configuration. 
This enables information from the card-play phase to bias state
sampling. While this method is shown to be effective, the independence
assumption does not align with the fact that for a given configuration, the
probability that a given card is present is highly dependent on the presence
of other cards. For instance, their approach cannot capture situations in
which a player's actions indicate that their hand likely contains
either the clubs jack or the spades jack, but not both.  The Policy Inference
approach presented in this paper captures this context by estimating a state's
probability based on the exact card configuration of that state.  In this way,
more precise inference is possible even though both techniques use the same 
body of data. 
The major upside of the Card Location
Inference (CLI) is that it runs much faster than the Policy Inference. 

In other domains, Richards and Amir \cite{richards2007opponent} model the 
opponent's policy using a static evaluation technique and then perform
inference on the opponent's remaining tiles given their most recent
move in Scrabble.
Baier et al.~\cite{baier2018emulating} leverage policies trained from 
supervised human data to bias MCTS results; this is similar to our approach in
that it uses human data to train an opponent model, but different because their
model is not used to infer opponent hidden information.
Sturtevant and Bowling \cite{sturtevant2006robust} build a generalized model of
the opponent from a set of candidate player strategies. 
Our use of aggregated human data could be viewed as a general model that
captures common action preferences from a large, diverse player base.

\subsection{Skat}

Our application domain for this paper is the game of Skat.  It is a 3-player
trick-taking card game that originates in Germany in the 1800s and is
played competitively around the world. Skat is played using a 32 card deck
where cards 2 through 6 from each suit are removed from the standard 52 card
deck.

Play starts after each player is dealt 10 cards; the two that remain are called 
the ``skat'', and are placed face down in the middle.
After observing their cards, players engage in the bidding phase to see which
of them play against the other two in the subsequent cardplay phase.
In the bidding phase, players alternate making successively higher bids based
on their hand and the highest-valued game they believe they could win. 
This value is dependent on the game type (see Table~\ref{tab:gametype}) 
and a multiplier that is based on the cards in the player's own hand and the 
outcome of the game (see Table~\ref{tab:modifiers}).

\begin{table}[t]
  \caption{Game Type Description}
  \label{tab:gametype}
  {
  \small
  \begin{tabular}{cccc}
        & Base  &        & Soloist Win\\ 
   Type & Value & Trumps & Condition\\
    \hline
    Diamonds & 9 & Jacks and Diamonds & $\ge$ 61 card points\\
    Hearts & 10 & Jacks and Hearts & $\ge$ 61 card points\\
    Spades & 11 & Jacks and Spades & $\ge$ 61 card points\\
    Clubs & 12 & Jacks and Clubs & $\ge$ 61 card points\\
    Grand & 24 & Jacks & $\ge$ 61 card points\\
    Null & 23 & No trump & losing all tricks\\
  \end{tabular}
  }
\bigskip
  \caption{Game Type Modifiers}
  \label{tab:modifiers}
  {\small
  \begin{tabular}{cl}
    Modifier & Description \\
    \hline
    Schneider & $\ge$90 card points for soloist\\
    Schwarz & soloist wins all tricks\\
    Schneider Announced & soloist loses if card points $<90$\\
    Schwarz Announced & soloist loses if opponents win a trick\\
    Hand & soloist does not pick up the skat\\
    Ouvert & soloist plays with hand exposed\\
  \end{tabular}
  }
  \vspace{-0.5cm}
\end{table}

Once the highest bidder is determined, that player has the option of picking up
the skat and discarding any two cards from their hand.
The same player declares a game type, which determines the specific rules 
used in the upcoming cardplay phase --- including the win condition and which
suit will be trump.

As in other trick-taking card games, the cardplay phase revolves around
winning tricks. Tricks start with the trick leader playing a card and proceed
in clockwise order. Players must play a card from the same suit as the card
that was initially played by the leader if they have one.  Otherwise, any card
can be played.  After every player has played a card, the highest ranked card
of either the led suit or the trump suit (if a trump was played) wins.

All game types involve the ``soloist'' (the player who won the bidding)
playing against a team of ``defenders'' (the other two players).  In suit and
grand games, both parties receive points for winning tricks containing certain
cards.  The soloist is required to amass at least 61 out of the possible 120
points to win the game.  In null games, the soloist must lose every trick to
win.  The soloist's score is either increased by the game value if the game
was won, or decreased by double the game value if it was lost.  Players play a
sequence of 36 of such hands and keep a tally of the score over all hands to
determine the overall winner in the competitive setting.

\section{Inference} \label{sec:inference}

To determine the probability of a given state $s$ in an information set $I$,
we need to calculate its reach probability $\eta$. If we can perfectly
determine the probability of each action that leads to this state, we can
simply multiply all the probabilities together and get $\eta$. Each $s$ in $I$
has a unique history $h$, the sequence of all previous $s$ and $a$ that lead
to it. $h \cdot a$ represents the history appended with the action action
taken at that state. Thus, there is a subset of $h$, containing all the $h
\cdot a$ for a given $s$. Formally:
 
\begin{equation}
  \eta(s|I) = \prod_{h \cdot a \sqsubseteq s} {\pi(h,a)} 
\label{eq:reach_prob}
\end{equation}

For trick-taking card games, the actions are either taken by the world (chance
nodes in dealing), other players' actions, and our actions. Transition
probabilities of chance nodes can be directly computed since these are only
related to dealing, and the probability of our actions can be taken as 1 since
we chose actions that lead to the given state with full knowledge of our own
policy. This leaves us with determining the move probability of the other
players. If we have access to the other players' policies, we can use
Equation~(\ref{eq:reach_prob}) to perfectly determine the probability we are
in a given state within the information set. If we repeat this process for all
states within the information set, we can calculate the probability
distribution across states. If we can perfectly evaluate the value of all the
state-action pairs, we can select the action that maximizes this expected
value which provides an optimal solution.

There are two main issues with this approach. The first is that we either do
not have access to the other players' policies, or they are expensive to compute. This makes opponent/partner modelling necessary, in which we assume
a computationally inexpensive model of the other players, and use them to estimate the reach probability
of the state. The second problem is that the number of states in the
information set can be quite large. To get around this, we can sample the
worlds and normalize the distribution over the subset of states. Because in
Skat the information set size for a player prior to card-play can consist of
up to 2.8 billion states we employed sampling.  We use
Algorithm~\ref{alg:estimate} to estimate a state's relative reach
probability. When we do not sample, this becomes an estimate of the states
true reach probability.

\begin{algorithm}[t]
  \DontPrintSemicolon
  \SetKwFunction{FMain}{EstimateDist}
  \SetKwFunction{sample}{SampleSubset}{}
  \SetKwFunction{normalize}{Normalize}{}
  \SetKwFunction{hist}{StateActionHistory}{}
  \SetKwProg{Fn}{}{}{}
  \Fn{\FMain{\textrm{InfoSet} $I$, \textrm{int} $k$, \textrm{OppModel} $\pi$}}
  {
    $S \leftarrow \sample{I, k}$ \;
    \For{$s \in S$}
    {
      $\eta(s) \leftarrow 1$ \;
      \For{$h,a \in \hist{$I$}$}
      {
        $\eta(s) \leftarrow \eta(s) * \pi(h, a)$
      }
    }
    \Return \normalize{$\eta$}
  }
  \vspace{0.5cm}
\caption{Estimate the state distribution of an information set 
  given an opponent model and the actions taken so far.}
\label{alg:estimate}
\end{algorithm}

While we have access to the policy of the current strongest Skat bot, using
its policy directly would be computationally intractable because it uses an
expensive search based method that also performs inference. Also, it is the
goal of this research to develop robust inference that is not based upon the
play of a single player. Thus, we decided to use policies learned directly from
a large pool of human players. These policies are parameterized by deep neural
networks trained on human games \cite{rebstock2019learning}. The features
for the pre-cardplay networks are a lossless one-hot encoding of the game
state while considerable feature engineering was necessitated for the cardplay
networks. Separate networks were trained for each distinct decision point in
the pre-cardplay section, and for each game type for the cardplay
networks. More details on the training and the dataset can be found in 
\cite{rebstock2019learning}.

The decision points in pre-cardplay are bidding, picking up or declaring a
hand game, choosing the discard, and declaring the game. The decision points in
the cardplay section are every time a player chooses what card to play. While
inference would be useful for decision-making in the pre-cardplay section, we
are only applying it to cardplay in this paper. As such, we can abstract the
bidding decisions into: the maximum bids of the bidder and answerer in the
bid/answer phase, and the maximum bids of bidder and answerer in the
continue/answer phase. For these maximum bid decision points, we only observe
the maximum bid if the player passes. For the cases in which the intent of
maximum bid is hidden, the probability attached to that decision point is the
sum of all actions that would have resulted in the same observation, namely
the probability of all maximum bids greater than the pass bid. The remaining
player decision points are pickup or declare a hand game, discard and declare, and
which card to play. As these are not abstracted actions, the probability of the
move given the state can be determined directly from the appropriate trained
network.

The current state-of-the-art Skat bot, Kermit, uses search-based evaluation
that samples card configurations.
A card configuration is the exact
location of all cards, and thus doesn't take into account which cards where
originally present in the soloist's hand prior to picking up the
skat. Depending on the game context, there are either 1 or 66 (12 choose 2)
states that correspond to a card configuration during the cardplay phase. Two
variants of inference were explored. The first variant samples card configurations. Decision points are ignored for inference 
if there are multiple states with the same card configuration but different features (input to the network). 
The second variant samples states directly, thus avoiding this issue. For our implementation, the need to
distinguish states that share a configuration only occurs when a player does
inference on the soloists actions prior to picking up the two hidden cards in
the skat. Sampling card configurations will be treated as the default approach
for PI. When states are sampled instead, the inference will be labelled PIF,
for Policy Inference Full.

\section{Experiments} \label{sec:exp}

In this section, we test the quality of the inference directly and indirectly
through the players performance in a tournament setup. The baseline players
are all versions of Kermit, with the only difference being the inference
module used. These inference modules are the original Kermit Inference (KI)\cite{buro2009improving},
card-location inference (CLI)\cite{solinas2019improving}, and no inference (NI).

\subsection{Direct Inference Evaluation}

To measure the inference quality directly, we measured the True State Sampling
Ratio (TSSR) \cite{solinas2019improving} for each main game type, separately 
for defender and soloist.
TSSR measures how many times more likely the true state will be sampled than
uniform random.

\begin{equation} \label{eq:tssr}
  TSSR = \eta(s^*|I)~/~(1/|I|) = \eta (s^*|I) \cdot |I|
\end{equation}
$\eta(s^*|I)$ is the probability that the true state is selected given the 
information set $I$, and $|I|$ is the number of possible states. Since the 
state evaluator of Kermit does not distinguish between states within the same card 
configuration, we will slightly change the definition to measure how many more 
times likely the card configuration (world) will be sampled than uniform random. 

Since the players tested use a sampling procedure when the number of worlds is
too large, the $TSSR$ value cannot be easily computed directly as this would
require all the world probabilities to be determined. We therefore estimate it
empirically. Since sampling was performed without replacement, we use the
given inference method to evaluate $\eta$ given that the true world was
sampled $k$ times. We can combine these values to get the combined probability
that the true world is sampled:

\begin{equation} \label{eq:sampled}
  TSSR = |I| \cdot \Sigma_k{BinDist(k,p) \cdot k \cdot \eta(s^*|I,k)} 
\end{equation}

where $BinDist$ is the probability mass of the binomial distribution with $k$
successes and probability of sampling the true world $p$ which is
$1/|I|$. Terms of the summation were only evaluated if the $BinDist(k,p)$
value were significant, which we cautiously thresholded at $10^{-7}$.

When the number of worlds is less than the set threshold parameter specific to
the player, we sample all worlds and can directly compute the value. For the
sake of the TSSR experiments, null games were further subdivided into the two
main variants, null and null ouvert. Null ouvert is played with an open hand
for the soloist, thus making the inference quite different from that of
regular null games from the perspective of the defenders. For each game in the
respective test set, the TSSR value was calculated for each move for the
soloist, and one of the defenders. The test set was taken from the human data,
and was not used in training. The number of games in the test sets were 4,000
for grand and suit, 3,800 for null, and 13,000 for nul ouvert.
\begin{figure*}
	\centering
  \begin{subfigure}{0.44\textwidth}
    \center{\textbf{Soloist}}
  \end{subfigure}
  \begin{subfigure}{0.44\textwidth}
    \center{\textbf{Defender}}
  \end{subfigure}
	\begin{subfigure}{\textwidth}
	\centering
		\includegraphics[height=0.28\textwidth]{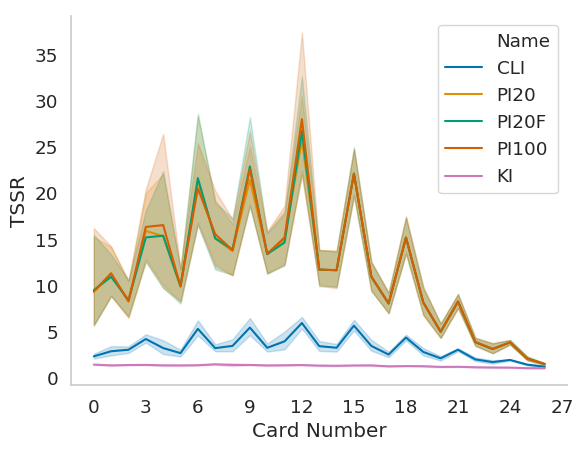}\hspace{4em}
		\includegraphics[height=0.28\textwidth]{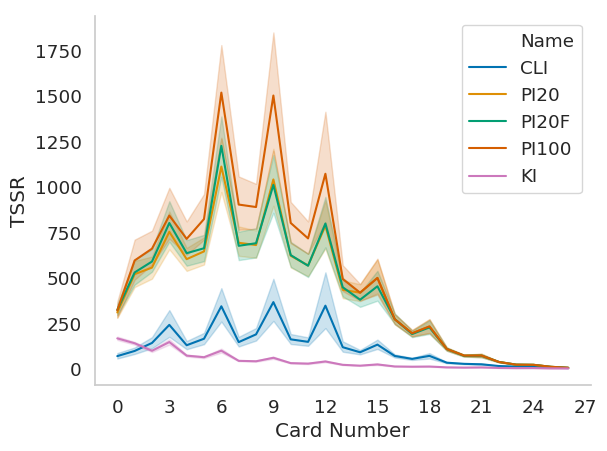}
		\caption{Grand}
		\label{fig:grand-tssr}
	\end{subfigure}
	\par
	\begin{subfigure}{\textwidth}
	\centering
		\includegraphics[height=0.28\textwidth]{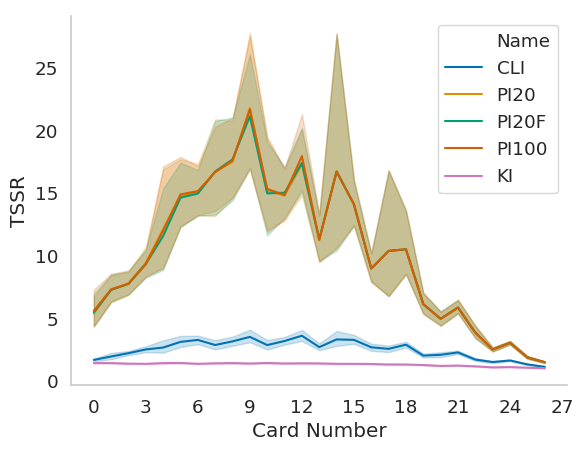}\hspace{4em}
		\includegraphics[height=0.28\textwidth]{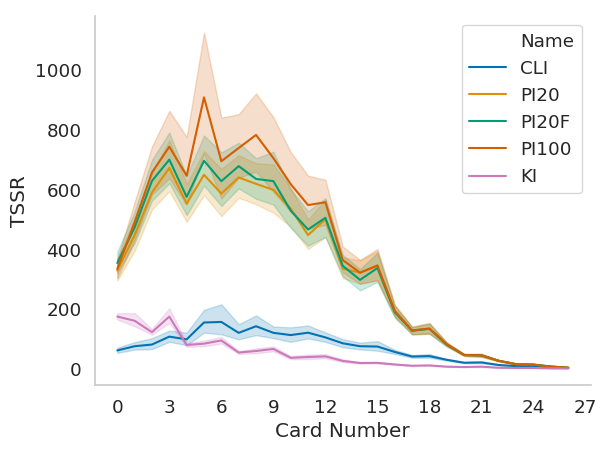}
		\caption{Suit}
    \label{fig:suit-tssr}
	\end{subfigure}
	\begin{subfigure}{\textwidth}
	\centering
		\includegraphics[height=0.28\textwidth]{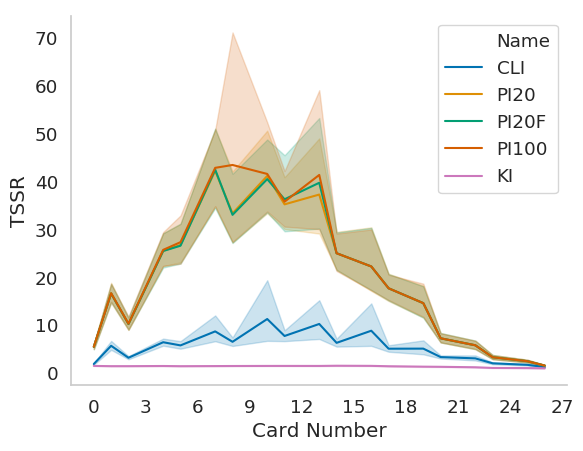}\hspace{4em}
		\includegraphics[height=0.28\textwidth]{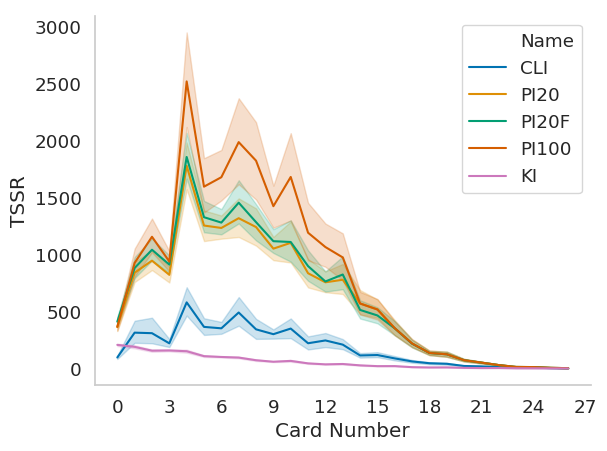}
		\caption{Null}
    \label{fig:null-tssr}
	\end{subfigure}
	\begin{subfigure}{\textwidth}
		\centering
		\includegraphics[height=0.28\textwidth]{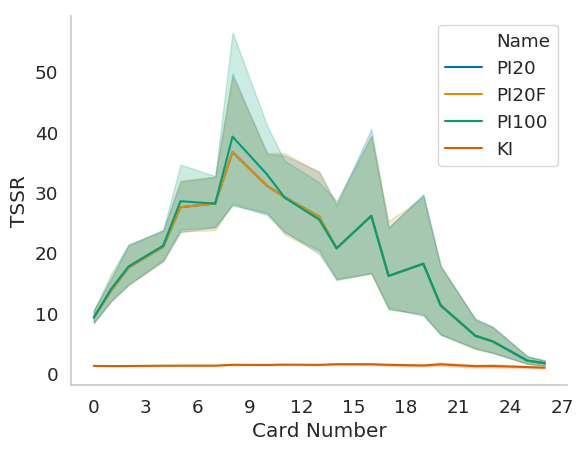}\hspace{4em}
		\includegraphics[height=0.28\textwidth]{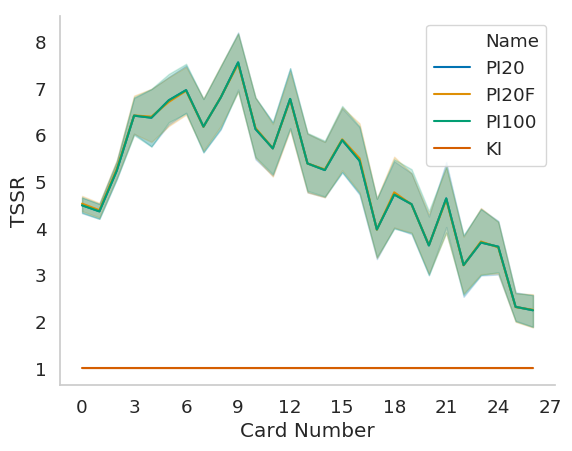}
		\caption{Null Ouvert}
    \label{fig:null-ouvert-tssr}
	\end{subfigure}
  \caption{Average \textit{TSSR} after \textit{Card Number} cards have been
  played. Data is separated by game type and whether the player to move is the
  soloist (left) or a defender (right).}
  \label{fig:tssr}
\end{figure*}

Figure \ref{fig:tssr} shows the average TSSR metric after varying number of 
cards have been revealed. The inference variants tested are PI20, PIF20, PI100, 
CLI, and KI. NI was not tested because it will always have a value of 1. PI20 and 
PI100 sample 20,000 and 100,000 card configurations 
respectively, while PIF20 samples 20,000 states. CLI samples 500,000 card 
configurations, and KI samples 3200 card configurations in soloist, and a 
varying number in defense. CLI inference was not implemented for null ouvert. 

TSSR is higher on defense, with the exception of null ouvert. This is likely
due to there being many more possible worlds in the
defenders information set because of the hidden cards in the skat. Also, the
defender can use the declaration of the soloist for inference, which is a 
powerful indicator of the soloist's hidden cards. Null ouvert does
not follow this trend because there are only 66 possible worlds at most in
defense while there are 184,756 for the soloist. This allows for higher TSSR
values for the soloist.

PI20, PIF20, and PI100 all achieve significantly higher TSSR values than the
other methods, across all game-types and roles. KI performs better than CLI at
the beginning of games, but surpasses KI once more cards are played.

PI100 appears to consistently perform better in defense than the other Policy
Inference variants, while PIF20 appears to perform slightly better than PI20
in the first half of defender games, but not significantly so. All TSSR values
trend down to 1 at the endgame, as the number of possible worlds approaches 1.

One common feature across all games is the spiking of TSSR values, which is
best exemplified in suit games. The spiking is consistent between players
within the same game type and role graph. However, between graphs it is not
consistently occurring at the same number of cards played.
We do not see an obvious reason for this. However,
these tests were done on human games and thus we are not controlling for
inherent biases in the distribution. Further investigation is needed to
determine why these spikes occur.

It is clear from these results that the
Policy Inference approach provides larger TSSR values, since the error envelopes are completely separated in the figure. It also should be noted that perfect inference would
not result in the upper bound TSSR value which is equal to the number of
worlds. Even with perfect knowledge of the opponents' policies, uncertainty is
inherent and thus a player with perfect TSSR value is not possible.

\subsection{Cardplay Tournament}

To test the performance of PI in cardplay, we played 5,000
matches for each of suit, grand, and null games against baseline players in a
pairwise setup. Only the cardplay phase of the game is played, while the
bidding and declaration is taken directly from the human data-set. These games
were held out from the policy training set. In a match, each player will play
as soloist against two copies of the opponent, as well as against two copies
of itself. The baseline players are all versions of Kermit,
with the only difference being the inference module used. These inference
modules are KI, CLI, with the addition of no inference (NI). This experiment
is designed to see if (a) the performance of the player improves as measured
by its play against opponents and (b) to determine the extent to which the
defender and soloist performance is responsible for this difference.

For each match-up we report the average tournament points per game (TP/G) for
the games in which the players played against each-other. The games in which
the player played against a copy of itself were used to determine the
difference in the effectiveness of the defenders and soloists.

$AvBB$ denotes a match-up in which the soloist is of type $A$ while the
defenders are both of type $B$. The value of the game $AvBB$ is in terms of
the soloist score, therefore it is the sum of the soloist's score and the
negation of the defenders' score. In this notation, the performance of player $A$ relative to player $B$
is given as
\begin{equation}
\Delta TP/G = [AvBB - BvAA]/3
\end{equation}

The value is divided by 3 since it is enforced that a player is soloist 1/3
the time in the tournament setup. To directly compare the performance of the
defenders, we can measure the performance difference between scenarios where
the only difference is the change in defenders.

\begin{equation}
\Delta Def/G = [(AvBB + BvBB) - (AvAA + BvAA)]/6
\end{equation}

A positive value for $\Delta Sol/G$ indicates $A$ performs better than $B$ in defense.
The same concept can also be applied to directly compare the efficacy of
the soloist.

\begin{equation}
\Delta Sol/G = [(AvAA - BvAA) + (AvBB - BvBB)]/6
\end{equation}

A positive value for $\Delta Sol/G$ indicates $A$ performs better than $B$ as the soloist.

\begin{table*}[t]
  \caption{Tournament results for each game type. Shown are average tournament
    scores per game for players NI (No Inference), CLI (Card-Location
    Inference), PI (Policy Inference), and KI
    (Kermit's Inference) which were obtained by playing 5,000 matches against
    each other, each consisting of two games with soloist/defender roles
    reversed. The component of $\Delta$TP attributed to Def and Sol is also 
    indicated}
  \label{tab:results}
  \setlength{\tabcolsep}{3pt}
  \begin{center}
    {\small\renewcommand*{\arraystretch}{1.2}
      \begin{tabular}{|c||c|c|c|c|c|c|c|c|c|c|c|c|}
      \hline
Game Type & \multicolumn{4}{c|}{Suit} & \multicolumn{4}{c|}{Grand} & \multicolumn{4}{c|}{Null}\\\hline
Matchup & TP & $\Delta$ TP & $\Delta$ Def & $\Delta$ Sol & TP & $\Delta$ TP & $\Delta$ Def & $\Delta$ Sol & TP & $\Delta$ TP & $\Delta$ Def & $\Delta$ Sol\\\hline\hline
KI : CLI & 17.62 : 20.81 & -3.19 &-2.76 &0.42$^*$ &37.02 : 38.98 & -1.96 &-1.85 &0.11$^*$ &17.22 : 19.83 & -2.61 &-2.70 &-0.09$^*$\\\hline
KI : PI & 16.48 : 21.61 & -5.13 &-4.22 &0.91 &36.44 : 39.56 & -3.12 &-2.30 &0.83 &17.29 : 19.66 & -2.37 &-2.93 &-0.56$^*$\\\hline
NI : CLI & 16.14 : 24.12 & -7.98 &-7.36 &0.62$^*$ &36.56 : 40.45 & -3.89 &-3.44 &0.46$^*$ &16.01 : 22.46 & -6.45 &-6.55 &-0.10$^*$\\\hline
PI : CLI & 19.50 : 17.18 & 2.32 &1.64 &-0.68$^*$ &37.87 : 37.23 & 0.64$^*$ &0.19$^*$ &-0.45$^*$ &18.84 : 17.27 & 1.57 &1.14 &-0.43$^*$\\\hline
KI : NI & 23.29 : 18.64 & 4.65 &4.53 &-0.12$^*$ &39.71 : 38.36 & 1.35 &1.30 &-0.05$^*$ &21.65 : 17.81 & 3.84 &3.85 &0.01$^*$\\\hline
NI : PI & 14.59 : 25.02 & -10.43 &-9.07 &1.36 &36.46 : 40.01 & -3.55 &-3.14 &-0.42$^*$ &15.77 : 22.10 & -6.33 &-6.84 &0.51$^*$\\\hline
  \end{tabular}}
  \end{center}
\end{table*}

The results for the tournament match-ups are shown in
Table~\ref{tab:results}. All $\Delta$ values reported have a $^*$ attached if
they are not found to be significant at a p value of 0.05 when a paired T-Test
was performed.  The general trend is that PI performs the best, followed by
CLI, then KI, then NI. This fits with the expectation that better TSSR values
seen in Figure~\ref{fig:tssr} would translate into stronger game
performance. Another interesting result is that the majority of the performance 
gain seems to be from the defenders, as demonstrated by the
$\Delta Def$ values being consistently larger than the $\Delta Sol$
values. The most interesting match-up is PI : CLI since it roots the previous
state-of-the-art skat inference against the new policy-based method. PI
outperforms CLI by 2.32, 0.64, and 1.57 TP/G in suit, grand, and null,
respectively. The grand result did not provide statistical significance.

The major drawback of the PI inference is the runtime. When combined with evaluation, PI20 takes roughly 5 times longer to make a move than CLI.

Further experiments were conducted to test the effect of increasing the number
of sampled worlds to 100,000 (PI100) and sampling states instead of card
configurations (PIF20). In addition, a cheating version of Kermit was
introduced (C) in which it places all probability on the true state. All
programs were tested against CLI with only the mirrored adversarial setup
used. The rest of the experimental setup was identical.

\begin{table*}[t]
  \caption{Tournament results for each game type. Shown are average tournament
    scores per game for players CLI (Card-Location
    Inference), PI20 (Policy Inference with 20,000 card configurations sampled), 
    PIF20 (Policy Inference with 20,000 states sampled), PI20 (Policy Inference 
    with 100,000 card configurations sampled), and C (Cheating Inference) which 
    were obtained by playing 5,000 matches against
    each other, each consisting of two games with soloist/defender roles
    reversed.}
  \label{tab:results2}
  \setlength{\tabcolsep}{3pt}
  \begin{center}
    {\small\renewcommand*{\arraystretch}{1.2}
      \begin{tabular}{|c||c|c|c|c|c|c|}
      \hline
Game Type & \multicolumn{2}{c|}{Suit} & \multicolumn{2}{c|}{Grand} & \multicolumn{2}{c|}{Null}\\\hline
Matchup & TP & $\Delta$ TP  & TP & $\Delta$ TP & TP & $\Delta$ TP\\\hline\hline
PI : CLI & 19.50 : 17.18 & 2.32 & 37.87 : 37.23 & 0.64$^*$ & 18.84 : 17.27 & 1.57 \\\hline
PIF20 : CLI & 19.30 : 17.65 & 1.65 & 37.67 : 37.20 & 0.47$^*$ & 18.01 : 17.66 & 0.35$^*$ \\\hline
PI100 : CLI & 19.55 : 16.69 & 2.86 & 38.19 : 36.09 & 2.10 & 18.13 : 17.10 & 1.03 \\\hline
C : CLI & 14.75 : 18.00 & -3.25 & 29.97 : 38.46 & -8.49 & 20.80 : 10.99 & 9.82 \\
\hline
PI : CLI (6way) & 19.06 : 17.82 & 1.24 & 37.64 : 37.27 & 0.38* & 18.31 : 17.74 & 0.57 \\
\hline
  \end{tabular}}
  \end{center}
\end{table*}

The results in Table~\ref{tab:results2} indicate that PI100 performs stronger
than the other PI variants in suit and grand, when playing against CLI, however, only the grand result is significant. The opposite is true for null, in which PI20 performs
strongest out of the PI variants. 
This result contradicts the idea that a higher TSSR value corresponds to 
better cardplay performance.
Cheating inference performs worse than CLI in all but
null games. This is interesting because it puts all the probability mass on
the true world, but still plays worse than a player that is not cheating. This
result in conjunction with the worse null game score for PI100 indicates that
further investigation into the exact role inference quality has within the
context of PIMC is required. Also, PI20 outperforms PIF20 over all game
types, showing that there can be benefits to sampling card configurations
instead of states when there is a limited sampling budget.

One further experiment was performed to determine whether performance gains
would be present with mixed defenders. This is interesting since it is possible the gain would only be present if the partner's inference was compatible with their own. For the sake of time, this was only done for the CLI and PI20 matchup. The added
arrangements are $AvAB$, $AvBA$, $BvAB$, and $BvBA$.  With these added, we now
have six games for each tournament match. The results for this match-up are included in
Table~\ref{tab:results2}. PI is consistently stronger than CLI (the grand
result is not significant), but the effect size is smaller. 
This is expected because PI is now defending against PI in the mixed setup 
games. To further analyze the the relative effectiveness of
the players as soloist against only the mixed team defenders, we can
calculate:

\begin{equation}
\Delta Sol = [(AvAB - BvAB) + (AvBA - BvBA)]/6
\end{equation}

$\Delta Def_B$ measures the difference in effectiveness of a mixed defense ($A$ and $B$) and a pure defense of $A$'s. It is calculated by averaging the effect of 
swapping in player $B$ into defense for all match-ups that included two $A$'s 
on defense. The reverse can be done to find the
effect of swapping in $A$ to form a mixed defense. A positive value for $\Delta
Sol$ means that PI is more effective than CLI as soloist in the mixed
setting. A positive value for $\Delta Def_{PI}$ means defense improved when it
was added, and same for $\Delta Def_{CLI}$.

\begin{table*}[t]
  \caption{Tournament results for each game type in the 6-way match between CLI and PI20. 5,000 matches were played for each game type. }
  \label{tab:results3}
  \setlength{\tabcolsep}{3pt}
  \begin{center}
    {\small\renewcommand*{\arraystretch}{1.2}
      \begin{tabular}{|c||c|c|c|}
      \hline
Game Type & $\Delta Sol$ & $\Delta Def_{CLI}$ & $\Delta Def_{PI}$ \\\hline\hline
Suit &  1.00 & -1.06 & 0.59* \\\hline
Grand & 0.38* & -0.02* & 0.16*\\\hline
Null & 0.18* & -0.59* & 0.55*\\\hline
  \end{tabular}}
  \end{center}
\end{table*}

While all the values in Table~\ref{tab:results3} show the same trends of PI
performing better on defense and soloist across all game types, the effect is
only statistically significant for $\Delta Sol$ and $\Delta Def_{CLI}$ in the
suit games. These tests were done using pairwise TTests with a significance threshold of p=0.05.

\section{Conclusion} \label{sec:concl}

Policy Inference (PI) appears to provide much stronger inference than its
predecessors, namely Kermit Inference (KI) and Card Location Inference (CLI)
as demonstrated by the TSSR value figures. Across the board, the higher TSSR
values translate into stronger game-play as demonstrated in card-play
tournament settings. PI20 outperforms CLI by 2.32, 0.64, and 1.57 TP/G in
suit, grand, and null games, respectively. Also, it seems that increasing the number of states sampled increases
the performance of PI, however, this did not translate into the null game
type. Further investigation into this null game result is needed. We would
expect that when substantially increasing the sampling threshold, the state
sampling employed by PIF20 would be more effective. But under this limited
sampling regimen, sampling card configurations is more effective than sampling 
states.

Future work related to inference in trick-taking card games should focus on
the relationship between opponent modelling and exploitability. In order to
investigate the robustness of our approach, we could try learning a best response to it.
Likewise, adjusting player models online could enable us to
better exploit our opponents and cooperate with teammates.

Another direction is to experiment with heuristics that allow our algorithm to
prune states that are highly unlikely and stop considering them altogether.
This could help us sample more of the states that are shown to be realistic
given our set of human games and possibly improve the performance of the
search.

\section*{Acknowledgment}

We acknowledge the support of the Natural Sciences and Engineering Research Council of Canada (NSERC).

Cette recherche a \'et\'e financ\'e par le Conseil de recherches en sciences naturelles et en g\'enie du Canada (CRSNG).

\bibliographystyle{IEEEtran}
\bibliography{refs}

\end{document}